\documentclass[11pt]{article}

\usepackage[preprint]{acl}

\usepackage{times}
\usepackage{latexsym}

\usepackage[T1]{fontenc}
\usepackage[utf8]{inputenc}

\usepackage{microtype}

\usepackage{inconsolata}

\usepackage{graphicx}

\usepackage{amsmath}
\usepackage{amssymb}
\usepackage{enumitem}
\usepackage{algorithm}
\usepackage{algpseudocode} % 或者 \usepackage{algorithmic}
\usepackage{amsthm}
\usepackage{booktabs}
\usepackage{multirow}

\newtheorem{remark}{Remark} % 这行代码定义了小写的 remark 环境

\title{Correlation-Aware Structured Pruning for Large Language Models}

\author{
  \textbf{Sicheng Xu\textsuperscript{1}},
  \textbf{Hao Shi\textsuperscript{1}\thanks{The first two authors contributed equally to this work.}},
  \textbf{Wei Zhang\textsuperscript{2}},
  \textbf{Haoran Pang\textsuperscript{1}},
  \textbf{Zhenyu Ming\textsuperscript{2}\thanks{Correspondence to: Zhenyu Ming <mingzhenyu1@huawei.com>.}},
\\
  \textbf{Hao Wu\textsuperscript{1}},
  \textbf{Zhongyi Huang\textsuperscript{1}},
  \textbf{Xin Yao\textsuperscript{2}},
  \textbf{Gong Zhang\textsuperscript{2}}
\\
\\
  \textsuperscript{1}Department of Mathematical Sciences, Tsinghua University \\
  \textsuperscript{2}Theory Lab, Central Research Institute, 2012 Labs, Huawei Tech. Co. Ltd.
}

\begin{document}
\maketitle
\begin{abstract}
\cite{le2025probe}
Structured pruning is a promising approach for reducing the substantial inference costs of Large Language Models (LLMs) while maintaining hardware efficiency. Many existing  methods assess the importance of prunable units (e.g., channels or heads) in isolation, implicitly assuming that pruning errors are additive.  This independence assumption is often invalidated by the non-orthogonality of model weights and strong correlations between unit activations, potentially leading to performance degradation. To address this, we propose a Correlation-Aware Structured Pruning  method. We formulate the pruning objective as a cardinality-constrained binary quadratic program that explicitly models cross-unit dependencies in the reconstruction error. Since this binary quadratic program is NP-hard and difficult to solve exactly, we develop a greedy interaction algorithm based on dependency-aware marginal costs to optimize unit selection. Furthermore, we incorporate a gradient-based strategy to achieve adaptive layer-wise sparsity allocation across the entire model. Extensive experiments on mainstream LLMs demonstrate that incorporating correlation information yields competitive accuracy-efficiency trade-offs compared to representative structured pruning baselines.
\end{abstract}

\section{Introduction}
Large language models (LLMs) have advanced rapidly in recent years, demonstrating strong capabilities in instruction following, reasoning, and code generation~\cite{brown2020language,chowdhery2023palm,touvron2023llama}.  {These performance gains come with} ever-increasing model sizes and inference-time resource demands. The large parameter count and activation footprint make LLM deployment challenging, imposing substantial latency, memory, and energy costs—especially in resource-constrained settings.

To reduce the cost of LLM inference, a variety of model compression techniques~\cite{zhu2024survey} have been explored, including pruning~\cite{ma2023llm,sun2024wanda}, quantization~\cite{frantar2023optq,xiao2023smoothquant}, and knowledge distillation~\cite{xu2024survey}. {Among these approaches, pruning stands out as a direct and effective strategy for model size reduction. It directly reduces model complexity by removing redundant weights or architectural units while preserving the original model structure.} {Pruning can be broadly divided into two categories:}  unstructured and structured pruning. Unstructured pruning,  {such as} SparseGPT~\cite{frantar2023sparsegpt} and Wanda~\cite{sun2024wanda}, removes individual weights and yields sparse weight matrices. Structured pruning ~\cite{ma2023llm,men2025shortgpt}, removes higher-level units such as attention heads, intermediate channels, or even entire layers, {yielding} a compact model  with {the original model's} dense-operator form. In this work, we focus on structured pruning, as its regular structure {better supports}  efficient execution on standard accelerator hardware and dense GEMM kernels~\cite{ma2023llm}.

{Recent advancements in structured pruning for LLMs have focused on the post-training regime, with several representative methods emerging.} LLM-Pruner~\cite{ma2023llm} ensures structural consistency by identifying coupled units via dependency graph analysis and pruning them based on gradient sensitivity. LoRAPrune~\cite{zhang2024loraprune} integrates pruning with Low-Rank Adaptation (LoRA) {to reduce gradient memory overhead}, utilizing LoRA updates as an efficient proxy for importance estimation. Distinct from these unified pipelines, LoRAP~\cite{li2024lorap} introduces a hybrid paradigm:  {attention weights suit low-rank decomposition, while FFN layers favor structured channel pruning.} Furthermore, recent studies have proposed novel metrics to capture the intrinsic redundancy in model weights---for instance, SlimGPT ~\cite{ling2024slimgpt} extends Optimal Brain Surgeon (OBS) to structured pruning, and FLAP~\cite{an2024fluctuation}  uses a fluctuation-based metric for feature stability to guide channel pruning. 

Most existing structured pruning approaches {evaluate} prunable units (e.g., attention heads or channels) in isolation, typically assigning a scalar importance score (or energy) to each unit based on local criteria~\cite{sun2024wanda,ma2023llm,li2024lorap}. This design {relies on the implicit assumption} that the pruning error induced by removing a set of units is approximately the sum of the errors incurred by removing each unit independently. However, { this independence assumption may not hold in LLMs}, as  the projection weights are generally non-orthogonal and intermediate activations exhibit strong correlations. {Since independent scoring fails to account for these cross-unit interactions, it may not fully capture the true impact of pruning. This motivates the need to develop structured pruning methods that account for cross-unit interactions.}

{Motivated by the need to account for unit dependencies}, we propose a Correlation-Aware Structured Pruning (CASP) method for LLMs in this work. {We start with a local pruning objective based on reconstruction error, which quantifies the impact of pruning a set of units on the output of a linear sub-layer.} By expanding the reconstruction error for pruning a set of units, {we find it naturally decomposes into two parts: a diagonal part for conventional per-unit importance and an off-diagonal part capturing cross-unit dependencies from weight alignment and activation correlations.}
 This decomposition motivates modeling pruning as a correlation-aware subset selection problem, rather than ranking units independently.
Building on this objective, we formulate a binary quadratic integer program, and develop a greedy interaction algorithm to generate high-quality masks at scale, {with the algorithm accounting for cross-unit dependencies.} To  {enhance} end-to-end performance, we further adopt a gradient-based layer-wise allocation strategy that adaptively adjusts compression rates according to the sensitivity of each layer. Experiments on mainstream LLMs demonstrate that incorporating {cross-unit dependency} information yields more robust pruning decisions and better accuracy--efficiency trade-offs than representative structured pruning baselines at the same sparsity level.

Our contributions are summarized as follows:
\begin{itemize}
    \item \textbf{Correlation-aware  {modeling}.}
    We derive an exact set-pruning reconstruction error and formulate structured pruning as a cardinality-constrained binary quadratic program that explicitly models channel dependencies.
    
    \item \textbf{Greedy Interaction Algorithm.}
    We design a greedy algorithm for the resulting quadratic program using interaction-aware marginal-cost updates, and adapt it to attention head pruning via block-wise aggregation.
    
    \item \textbf{Gradient-Based Sparsity  Allocation.}
    %To bridge local pruning decisions and end-to-end {performance}, we propose a gradient-based layer-wise compression allocation strategy that assigns higher compression to less sensitive layers and lower compression to more sensitive ones under a fixed global budget.
   {We propose a gradient-based layer-wise allocation strategy that allocates sparsity according to layer sensitivity under a fixed global sparsity level.}
\end{itemize}
\section{Preliminaries}
\label{sec:prelim}

\subsection{Transformer Mechanisms}
We use LLaMA-style decoder-only Transformers as an example and define the notation for a standard Transformer block,
which consists of a multi-head self-attention (MHA) sub-layer and a gated feed-forward network (FFN) sub-layer.
Let $X \in \mathbb{R}^{L \times d}$ denote the input activation, where $L$ and $d$ are the sequence and hidden dimensions, respectively.

Given $h$ attention heads and per-head dimension $d_h$ with $d = h d_h$, the $i$-th attention head applies linear projections to obtain query, key, and value representations:
\begin{equation*}
\label{eq:qkv_i}
Q^{(i)} = XW_{q}^{(i)},\quad K^{(i)} = XW_{k}^{(i)},\quad V^{(i)} = XW_{v}^{(i)},
\end{equation*}
where $W_{q}^{(i)}, W_{k}^{(i)}, W_{v}^{(i)} \in \mathbb{R}^{d \times d_h}$ are learnable projection matrices. 
The head output is computed by scaled dot-product attention:
\begin{equation*}
\label{eq:mha_head}
\mathrm{head}_i
=
\mathrm{Softmax}\!\left(\frac{Q^{(i)}{K^{(i)}}^\top}{\sqrt{d_h}}\right)\, V^{(i)}.
\end{equation*}
The multi-head attention concatenates all head outputs along the feature dimension and projects them back to $\mathbb{R}^{L \times d}$:
\begin{equation*}
\label{eq:mha}
\mathrm{MHA}(X)=\mathrm{Concat}(\mathrm{head}_1,\ldots,\mathrm{head}_h)\,W_o,
\end{equation*}
where $W_o \in \mathbb{R}^{d \times d}$ is the output projection.

The FFN expands $d$ to an intermediate width $d_m$ and projects back to $d$ through an element-wise gated activation:
\begin{equation*}
\label{eq:ffn}
\mathrm{FFN}(X)
=
\bigl( (X W_{\text{up}}) \odot \sigma(X W_{\text{gate}}) \bigr)\, W_{\text{down}},
\end{equation*}
where $W_{\text{up}}, W_{\text{gate}} \in \mathbb{R}^{d \times d_m}$ project $X$ into the intermediate space and $W_{\text{down}} \in \mathbb{R}^{d_m \times d}$ maps it back to the hidden dimension.
Here $\odot$ denotes element-wise multiplication, and $\sigma(\cdot)$ is the SiLU activation.

\subsection{Layer-Wise Pruning}
Structured pruning removes entire architectural units to obtain hardware-friendly acceleration.
For LLMs, it is commonly applied along four dimensions: attention heads, FFN intermediate neurons (channels),
embedding width, and model depth. In this work, we focus on pruning (i) attention heads in the MHA sub-layer and
(ii) intermediate channels in the gated FFN, {allowing non-uniform sparsity across layers}.

A unified view is to cast structured pruning as selecting a subset of rows in a linear map.
Let $Y \in \mathbb{R}^{L \times d_m}$ denote the input to a projection matrix $W \in \mathbb{R}^{d_m \times d}$,
so the dense output is $Z=YW$.
We introduce a binary mask $M \in \{0,1\}^{d_m \times d}$ (or row-wise mask) to zero out selected structures in $W$,
and define the reconstruction objective
$$
\min_{M}\ \big\|YW - Y\big(M\odot W\big)\big\|_2^2 .
$$
This objective instantiates to the two targets in this paper:(1) \textbf{Attention-head pruning.} Writing the MHA output as $O \!=\! HW_O$ (with $H$ the concatenation of head outputs),
pruning a head corresponds to masking the associated block of the output projection $W_O$, i.e., $Y \!=\! H$ and $W \!=\! W_O$.
(2) \textbf{FFN channel pruning.} For the gated FFN, the down-projection takes $Z\!=\!YW_{\text{down}}$ with
$Y\!=\!(XW_{\text{up}})\odot\sigma(XW_{\text{gate}})$; pruning an intermediate channel corresponds to masking the
corresponding row(s) in $W_{\text{down}}$, i.e., $Y$ is the intermediate activation and $W \!=\! W_{\text{down}}$.
Once heads/channels are selected for pruning, the corresponding parameters in the preceding projections are pruned accordingly (e.g., $W_q^{(i)}, W_k^{(i)}, W_v^{(i)}$ for attention heads, and $W_{\text{up}}, W_{\text{gate}}$ for FFN channels).

{Existing methods ~\cite{sun2024wanda,ma2023llm,li2024lorap} typically assign an importance score to each prunable unit independently (e.g., based on reconstruction error energy combining weight magnitude and activation energy) and retain units with the highest scores.}
%In practice, the above subset selection is combinatorial. A typical approach is to assign an energy (importance) score to each prunable unit and then prune those with the smallest scores. Let $w_j^\top$ denote the $j$-th row of $W$ and $y_j$ denote the corresponding $j$-th column of $Y$. If pruning the $j$-th unit induces the output change $\Delta_j = y_j w_j^\top$, then using the squared reconstruction error as the energy gives 
%\[
%E_j \triangleq \mathbb{E}\!\left[\|\Delta_j\|_F^2\right]
%= \|w_j\|_2^2\,\mathbb{E}\!\left[\|y_j\|_2^2\right].
%\]
%This score can be interpreted as combining the weight magnitude $\|w_j\|_2^2$ and the activation energy $\mathbb{E}[\|y_j\|_2^2]$, and pruning keeps units with the largest energies.

\subsection{Notations}
\label{sec:error_decomposition}

Building on the linear map $Z=YW$ introduced above, we decompose the output as $Z=\sum_{j=1}^{d_{m}}y_{j}w_{j}^{\top}$, {where $y_{j}$ denotes the $j$-th column of $Y$ and $w_{j}^{\top}$ denotes the $j$-th row of $W$.} Let $S \subseteq \{1, \dots, d_m\}$ be the set of $k$ pruned indices, {and the pruned output is then given by} $\hat{Z} = \sum_{j \notin S} y_j w_j^\top$. The reconstruction error is quantified via the expected squared Frobenius norm~\cite{ling2024slimgpt}:
\begin{equation}
\mathcal{L}(S)\!=\!\mathbb{E}\!\left[\!\|Z - \hat{Z} \|_F^2\!\right]
\!=\!\mathbb{E}\!\left[\left\|\sum_{i\in S} y_i w_i^\top\right\|_F^2\right].
\label{error}
\end{equation}

\section{Related Work}
\label{sec:related}

\paragraph{Pruning Granularity.} Model pruning approaches can be categorized by their granularity: unstructured pruning~\cite{frantar2023sparsegpt,sun2024wanda} removes individual weights, while structured pruning removes coherent architectural units (e.g., layers, blocks, attention heads,FFN channels) for hardware-friendly acceleration.
Structured pruning spans from coarse-grained layer removal~\cite{men2025shortgpt,sieberling2025evopress} to finer-grained channel and head-wise pruning~\cite{ma2023llm,guo2025slimllm}. Recent works focus on channel and head-wise pruning, which balances latency reduction and model capacity preservation.

%Model pruning approaches can be systematically categorized by their \emph{granularity}, which defines the structural level at which parameters are removed. At the finest granularity, unstructured pruning~\cite{frantar2023sparsegpt,sun2024wanda} eliminates individual weights independently. While this allows for high theoretical compression rates with minimal accuracy loss, the resulting irregular sparsity patterns are notoriously difficult to accelerate on commodity hardware, often requiring specialized kernels to yield actual speedups.To address these deployment challenges, structured pruning increases the granularity to remove coherent architectural units, ensuring the pruned model retains a dense operator form compatible with highly optimized GEMM kernels. This category spans a spectrum of coarse-to-fine structures: from removing entire transformer layers or blocks (coarse-grained), to pruning specific attention heads or intermediate FFN channels (fine-grained structured). While coarse-grained methods ~\cite{men2025shortgpt,sieberling2024evopress} offer substantial throughput improvements, they often incur significant accuracy degradation. Consequently, recent LLM pruning works~\cite{ma2023llm,guo2025slimllm} have largely converged on the channel and head-wise granularity. This middle ground strikes a favorable balance, providing meaningful latency reductions while preserving sufficient model capacity to maintain linguistic capabilities.

\paragraph{Pruning Criteria.} 

Pruning efficacy depends on the criterion used to rank units. Existing methods fall into three categories: (1) Magnitude and activation-based heuristics~\cite{sun2024wanda} utilize zero-order statistics for efficiency. (2) Gradient-based sensitivity methods~\cite{ma2023llm,zhang2024loraprune} estimate the impact of pruning on the global loss. (3) Reconstruction-based objectives focus on preserving layer-wise outputs~\cite{frantar2023sparsegpt,ling2024slimgpt,an2024fluctuation}. Most existing methods assess units independently, which may overlook cross-unit interactions prevalent in LLMs.
{Early work has explored interaction-aware formulations for pruning smaller-scale Transformer models~\cite{kwon2022fast}, which use Fisher approximations to model unit correlations. However, such curvature-based methods become computationally expensive and numerically unstable for modern LLMs due to the massive dimensionality of Transformer layers.}

\paragraph{Layerwise Allocation.}
To address varying sensitivities across Transformer layers, recent methods employ non-uniform allocation strategies instead of uniform pruning. SlimLLM~\cite{guo2025slimllm} adjusts per-layer pruning rates based on representation change, while LoRAP~\cite{li2024lorap} advocates for differentiated compression across sublayers. Additionally, SlimGPT~\cite{ling2024slimgpt} investigates incremental, non-uniform pruning ratios to mitigate performance degradation in high compression regimes. These works show that layer-wise allocation is critical, especially under high compression.

\section{Methodology}

\begin{figure*}[ht]
    \centering
    \includegraphics[width=\textwidth]{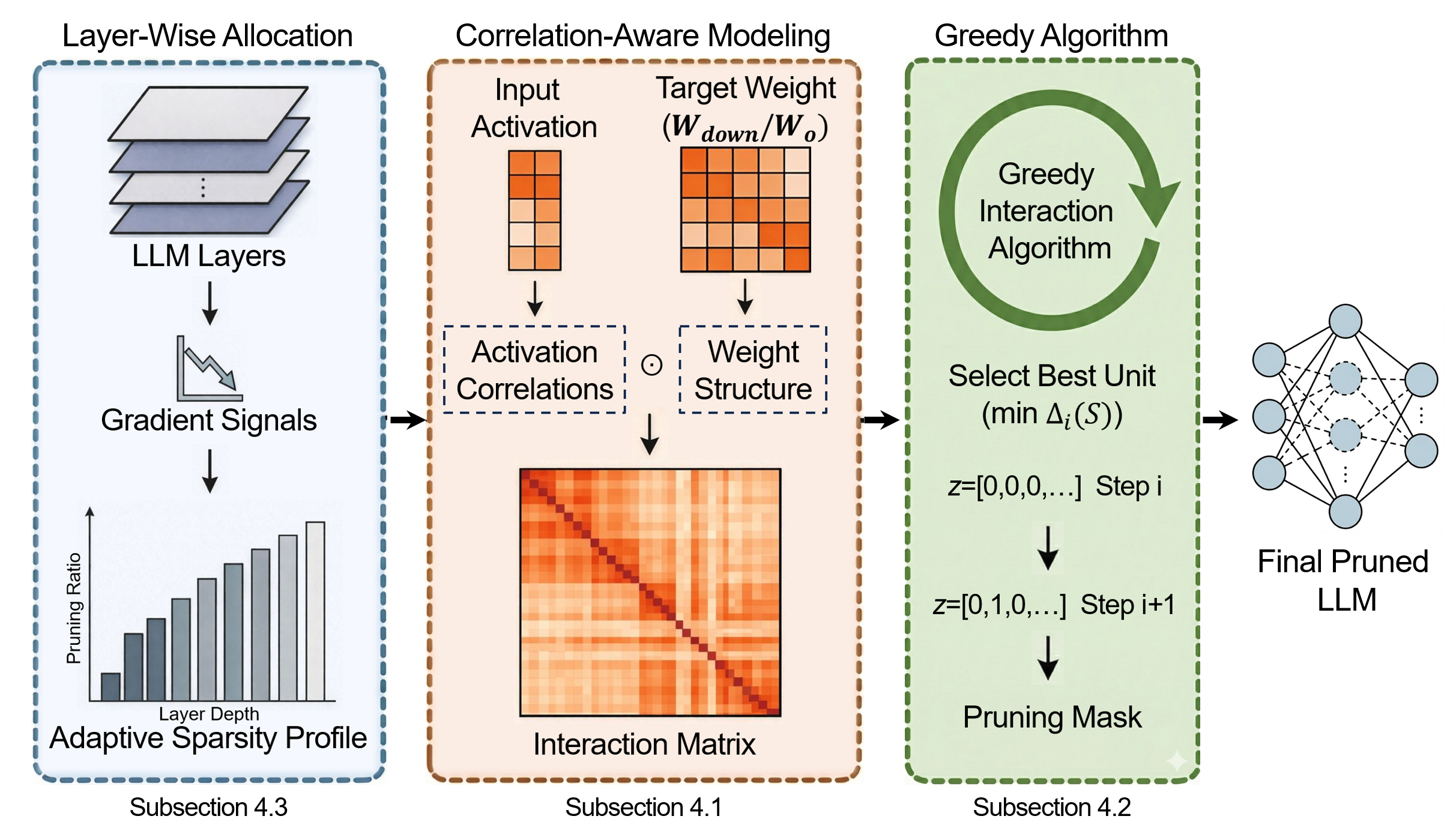}
    \caption{\textbf{The Correlation-Aware Structured Pruning (CASP) workflow.}
    Linear pipeline (left-right): (1) Layer-Wise Allocation generates a sparsity profile; (2) Correlation-Aware Modeling builds interaction matrix $Q$; (3) Greedy Algorithm selects optimal units to yield the pruned LLM.}
    \label{fig:pipeline}
\end{figure*}

To address the limitations of independent unit scoring, we propose the Correlation-Aware Structured Pruning method: {key components} include explicitly modeling cross-unit dependencies and designing a greedy interaction algorithm to handle the resulting optimization problem. Complementing these components, we incorporate a gradient-based layer-wise sparsity allocation strategy as a preprocessing step, which adaptively determines pruning ratios across layers. Together, these three components enable performance-preserving pruning of LLMs, with the full execution pipeline detailed in Figure \ref{fig:pipeline}.

\subsection{Correlation-Aware Modeling}
\label{sec:modeling}

To explicitly account for cross-unit dependencies in structured pruning, we formulate the pruning task as a constrained optimization problem:
\begin{equation}
\begin{aligned}
\min_{z\in\{0,1\}^{d_m}} \quad & z^\top Q z \\ \text{s.t.}\quad & \mathbf{1}^\top z = k,
\end{aligned}
\label{eq:qubo_card}\end{equation}
where $k$ denotes the number of units to be pruned.

This quadratic formulation is obtained by expanding the expected reconstruction error $\mathcal{L}(S)$ in Eq.~\eqref{error}. By expanding the squared Frobenius norm, we decompose the error into individual and interactive components:
\begin{equation}
\label{eq:exact_expand}
\begin{split}
\mathcal{L}(S) &= \sum_{i\in S}\sum_{j\in S}
\mathbb{E}\!\left[
\left\langle y_i w_i^\top,\; y_j w_j^\top \right\rangle_F
\right]\\
&= \underbrace{\sum_{i\in S}\|w_i\|_2^2 \cdot \mathbb{E}\!\left[\|y_i\|_2^2\right]}_{\text{diagonal / individual terms}} \\
&\quad + \underbrace{\sum_{\substack{i,j\in S\\ i\neq j}} (w_i^\top w_j)\cdot \mathbb{E}\!\left[y_i^\top y_j\right]}_{\text{off-diagonal / interaction terms}}.
\end{split}
\end{equation}

Let $z \in \{0,1\}^{d_m}$ denote the binary indicator for pruned units with budget $k$, corresponding to the index set $S=\{i: z_i=1\}$. We then define the interaction matrix $Q \in \mathbb{R}^{d_m \times d_m}$ via
\begin{equation}
Q_{ij}\triangleq (w_i^\top w_j)\cdot \mathbb{E}\!\left[y_i^\top y_j\right],
\end{equation}
which leads to the optimization problem formulated in Eq.~\eqref{eq:qubo_card}.

Computationally, $Q$ can be efficiently obtained via element-wise operations as $Q = G \odot C$, where $G = WW^\top$ captures the weight geometry, and $C = \mathbb{E}[Y^\top Y]$ represents the activation covariance estimated over a small calibration set.

This interaction matrix $Q$ offers a precise physical interpretation of how pruning decisions impact model performance:
\begin{itemize}
    \item \textbf{Diagonal Terms ($Q_{ii}$):} These terms quantify the error induced by pruning unit $i$ in isolation, corresponding to standard independent importance scores found in existing literature.
    \item \textbf{Off-Diagonal Terms ($Q_{ij}, i \neq j$):} These terms measure the \textit{interaction} between units. A positive value ($Q_{ij} > 0$) indicates error amplification (pruning $i$ and $j$ together is more damaging than the sum of their individual removal). Conversely, a negative value ($Q_{ij} < 0$) indicates error cancellation, suggesting that the errors from removing these units partially offset each other, thereby reducing the overall pruning cost.
\end{itemize}

Most existing structured pruning methods assess unit importance independently, which implicitly amounts to neglecting the off-diagonal terms of $Q$. This treatment relies on the assumption that different units are effectively decoupled, requiring the corresponding weight vectors to be orthogonal and hidden-state dimensions to be uncorrelated. However, these conditions are rarely satisfied in modern LLMs, where dense-layer weights are non-orthogonal and hidden states exhibit pronounced inter-dimensional correlations~\cite{ethayarajh2019contextual}. Consequently, explicitly considering the off-diagonal terms of $Q$ becomes essential. We further empirically validate the benefits of incorporating these cross-unit dependencies in  Section~\ref{sec:exp}.

The formulation in Eq.~\eqref{eq:qubo_card} corresponds to a $0$--$1$ quadratic program, which is NP-hard in general~\cite{garey2002computers}.  Given the massive dimensionality of LLM layers (where $d_m$ can exceed 10,000), obtaining an exact global solution is computationally intractable. {This necessitates the use of an efficient approximation strategy}, which we detail in the following subsection.

\subsection{Greedy Interaction Algorithm}
\label{sec:greedy_solver}

To solve the combinatorial optimization problem formulated in Eq. \ref{eq:qubo_card}, we start with a key observation: the objective function exhibits an incremental structure, where the marginal cost of pruning a unit can be exactly derived from the current pruned set. This observation naturally leads to a greedy strategy: we construct the pruning mask sequentially, selecting at each step the unit that minimizes the incremental reconstruction error. Unlike conventional greedy pruning that relies on fixed importance scores, the marginal cost in our strategy depends dynamically on previously selected units, enabling the algorithm to explicitly capture cross-unit dependencies.

%Although the global optimization is combinatorial, we observe that the objective admits an incremental structure: the marginal cost of pruning a unit can be derived exactly from the current pruned set. This suggests a natural greedy strategy that builds the pruning mask sequentially, selecting at each step the unit that induces the smallest increase in reconstruction error. Importantly, unlike conventional greedy pruning based on fixed importance scores, the marginal cost here depends dynamically on previously selected units, allowing the algorithm to explicitly account for cross-unit interactions.

Motivated by this observation, we propose an efficient Greedy Interaction Algorithm.
Starting from an empty pruned set $S = \emptyset$, we iteratively add the unit that minimizes the increase in reconstruction error.  For a candidate unit $i \notin S$ to the set, the marginal increase in the objective is:
\begin{equation}
\Delta_i(S)
\!=\!
\mathcal{L}(S\cup\{i\})\!-\!\mathcal{L}(S)
\!=\!
Q_{ii} \!+\! 2\sum_{j\in S} Q_{ij}.
\label{eq:marginal_cost}
\end{equation}

At each step $t$:
 \begin{itemize}[itemsep=2pt, topsep=0pt, parsep=0pt]
     \item Select the optimal unit $i^\star = \arg\min_{i \notin S} \Delta_i(S)$.
     \item Update $S \leftarrow S \cup \{i^\star\}$.
 \end{itemize}
This process repeats until the budget $|S| = k$ is satisfied. The complete procedure is formalized in Algorithm ~\ref{alg:greedy_interaction}.

\begin{algorithm}[t]
\caption{Greedy Interaction Search}
\label{alg:greedy_interaction}
\begin{algorithmic} % 新版通常不需要 [1] 参数，行号默认开启
    \State \textbf{Input:} interaction matrix $Q\in\mathbb{R}^{d_m\times d_m}$, prune count $k$
    \State $S\leftarrow \emptyset$; $u\leftarrow \mathbf{0}\in\mathbb{R}^{d_m}$
    \For{$t=1$ \textbf{to} $k$} % 注意：这里改为 \For，且不需要手动写 \ENDFOR
        \State Compute marginal costs $\Delta_i \leftarrow Q_{ii}+2u_i$ for all $i\notin S$
        \State Select $i^\star \leftarrow \arg\min_{i\notin S} \Delta_i$
        \State $S \leftarrow S \cup \{i^\star\}$
        \State $u \leftarrow u + Q_{:,i^\star}$
    \EndFor % 这一行其实可以省略，因为 \For 会自动生成结束行，但保留也无妨
    \State \textbf{Output:} pruned index set $S$ (indicator $z$)
\end{algorithmic}
\end{algorithm}

%To ensure computational efficiency, we avoid recomputing the summation in Eq.~\eqref{eq:marginal_cost} from scratch by maintaining  a running vector $u\in\mathbb{R}^{d_m}$:
%\begin{equation*}
%u_i = \sum_{j\in S} Q_{ij}.
%\end{equation*}
%Then $\Delta_i(S)=Q_{ii}+2u_i$ can be computed in $O(1)$ per candidate.
%After selecting $i^\star$, we update $u \leftarrow u + Q_{:,i^\star}$ in $O(d_m)$ time.
%Therefore, the total complexity per layer is $O(d_m\,k)$, which is feasible under moderate pruning ratios.

\begin{remark}
    We reduce the per-layer complexity from $O(k^2\,d_m)$ (naive computation) to $O(k\,d_m)$ by maintaining a running vector $u\in\mathbb{R}^{d_m}$ with $u_i = \sum_{j\in S} Q_{ij}$.
    This avoids recomputing the summation in Eq. ~\ref{eq:marginal_cost} from scratch, allowing $\Delta_i(S)=Q_{ii}+2u_i$ to be computed in $O(1)$ per candidate. After selecting $i^\star$, we update $u \leftarrow u + Q_{:,i^\star}$ in $O(d_m)$ time, thereby reducing the computational cost from quadratic to linear with respect to the cardinality constraint $k$.
\end{remark}

\begin{remark}
It is worth mentioning that the proposed interaction modeling introduces a quadratic cost in the module dimension, since constructing the interaction matrix 
$Q = (W^\top W) \odot \mathbb{E}[Y^\top Y]$ requires $O(d_m^2)$ computation. This cost is incurred only once during the offline pruning stage and is therefore not part of the forward computation of the deployed pruned model. Moreover, computing $Q$ mainly involves matrix multiplications and element-wise operations, which are highly parallelizable and can be efficiently accelerated on modern GPUs.
As shown in Appendix~\ref{appendix:time},  the overall pruning procedure remains  efficient in practice despite the quadratic interaction construction step.
\end{remark}

\begin{remark}
    Our method naturally extends to coarse-grained pruning, specifically for attention heads (where the pruning unit is a block of indices rather than a single channel). Let $\{\mathcal{B}_h\}_{h=1}^{H}$ be a partition of $\{1,\ldots,d_m\}$ where $\mathcal{B}_h$ contains the indices belonging to block $h$. By aggregating the interaction matrix $Q$ over these blocks, we obtain a reduced matrix $\widetilde{Q}\in\mathbb{R}^{H\times H}$:
    \begin{equation*}
    \label{eq:block_Q}
    \widetilde{Q}_{ab}
    \;\triangleq\;
    \sum_{i\in \mathcal{B}_a}\sum_{j\in \mathcal{B}_b} Q_{ij},
    \end{equation*}
    and the objective simplifies to $s^\top \widetilde{Q}\, s$. The same greedy interaction search can be applied to select $k$ heads, with marginal costs computed analogously.
\end{remark}

%Our framework also naturally extends to coarse-grained pruning units, specifically for attention heads. In this setting, the pruning unit is a block of indices rather than a single channel. Let $\{\mathcal{B}_h\}_{h=1}^{H}$ be a partition of $\{1,\ldots,d_m\}$ where $\mathcal{B}_h$ contains the indices belonging to head $h$. We introduce a head-level indicator $s\in\{0,1\}^{H}$, where $s_h=1$ implies pruning head $h$. By aggregating the interaction matrix $Q$ over these blocks, we obtain a reduced matrix $\widetilde{Q}\in\mathbb{R}^{H\times H}$:
%\begin{equation*}
%\label{eq:block_Q}
%\widetilde{Q}_{ab}
%\;\triangleq\;
%\sum_{i\in \mathcal{B}_a}\sum_{j\in \mathcal{B}_b} Q_{ij}.
%\end{equation*}
%The objective then simplifies to $s^\top \widetilde{Q}\, s$, allowing us to apply the same greedy interaction search on $\widetilde{Q}$ to select $k$ heads, with marginal costs computed analogously.

\begin{remark}
     Theoretically, problem \eqref{eq:qubo_card} can be approached by relaxing the binary constraint to $z \in [0,1]^{d_m}$. Since $Q$ is positive semi-definite, this relaxation constitutes a convex Quadratic Program (QP). However, direct optimization of this QP is computationally {expensive} for LLM layers due to their massive dimensionality, so we focus on the greedy algorithm in this work. Nevertheless, this relaxation still provides a rigorous theoretical lower bound for the discrete problem, allowing us to assess the optimality gap of our greedy solution, and can also serve as a baseline in experiments to validate the effectiveness of our method.
     We provide the detailed formulation and experimental comparisons in Appendix~\ref{appendix:continuous_relaxation} for interested readers.
\end{remark}

\subsection{Gradient-Based Compression Allocation}
\label{sec:grad_allocation}

This {subsection} introduces a gradient-based layer-wise compression allocation strategy as a preprocessing step for the pruning pipeline. This approach is designed to bridge the gap between minimizing local errors and optimizing end-to-end behavior. The process operates by first defining a layer sensitivity metric using gradient signals, and then normalizing these sensitivities to calculate layer-wise sparsity.

To bridge the gap between local pruning decisions and global model performance, we must assess how perturbations in each layer affect the final objective. Motivated by the first-order Taylor expansion, the change in the objective function $\mathcal{J}$ (e.g., KL divergence) induced by a perturbation of the $\ell$-th layer's output $\Delta Z_\ell$ can be approximated as:
\[
\mathcal{J}(Z_\ell + \Delta Z_\ell) - \mathcal{J}(Z_\ell) \approx \left\langle \frac{\partial \mathcal{J}}{\partial Z_\ell}, \Delta Z_\ell \right\rangle.
\]
Since the goal of local pruning is to {minimize} $\|\Delta Z_\ell\|_F^2$, we evaluate the end-to-end sensitivity of each layer using the expected  Frobenius norm of the gradient:
\[
S_\ell \triangleq \mathbb{E}\left[\left\|\frac{\partial \mathcal{J}}{\partial Z_\ell}\right\|_F^2\right].
\]
A larger $S_\ell$ indicates higher sensitivity, implying that a lower pruning ratio should be allocated to this layer.

We observe a general rule that the magnitude of {layer-wise} gradients decreases as the layer depth increases (experimental results are presented in the Appendix \ref{appedix:sensitivity}). This implies that deeper layers can tolerate higher compression, motivating us to adopt an increasing pruning ratio across layers, which is similar in spirit to the layer-wise allocation strategy in SlimGPT~\cite{ling2024slimgpt}.

To transform these sensitivity signals into concrete pruning ratios $\{r_\ell\}_{\ell=1}^{L}$ under a fixed global sparsity budget $r$, we employ a normalization process. First, to handle the wide dynamic range of sensitivity values, we apply a logarithmic transformation followed by min-max normalization:
\[
\tilde S_\ell = \log(S_\ell+\epsilon), \quad \hat S_\ell = \frac{\tilde S_\ell - \min_j \tilde S_j}{\max_j \tilde S_j - \min_j \tilde S_j},
\]
where $\epsilon$ is a small constant for numerical stability. 
{Subsequently}, we convert sensitivity to a compressibility score by inversion,
\[
c_\ell = (1-\hat S_\ell)^{\alpha},
\]
where $\alpha$ is a temperature parameter controlling how sharply compression is concentrated on less sensitive layers {(a larger $\alpha$ leads to more concentrated compression)}. Next, we scale the scores so that the mean sparsity matches the global budget:
\[
r_\ell = \frac{Lc_\ell}{\sum_{j=1}^{L}c_j} r.
\]
{This strategy enables the adaptive assignment of lower compression rates to more sensitive layers and higher rates to less sensitive layers.}

\begin{table*}[t]
\centering
\scriptsize
\setlength{\tabcolsep}{3.2pt}
\renewcommand{\arraystretch}{1.15}
\caption{Zero-shot performance of the compressed LLaMA2-7B. The average score is computed across seven datasets. The "\textbf{bolded}" represents the best result under the same pruning ratio.}
\resizebox{\textwidth}{!}{%
\begin{tabular}{cc|cc|ccccccccc|c}
\toprule
Ratio & Method
& WikiText2 $\downarrow$ & PTB $\downarrow$
& BoolQ $\uparrow$ & PIQA $\uparrow$ & HellaSwag $\uparrow$ & WinoGrande $\uparrow$
& ARC-e $\uparrow$ & ARC-c $\uparrow$ & OBQA $\uparrow$
& Average $\uparrow$ \\
\midrule

0\% & Ground Truth
& 6.9413 & 23.0732
& 0.8058 & 0.7617 & 0.5800 & 0.6835
& 0.7264 & 0.4172 & 0.3200
& 0.6135 \\
\midrule

\multirow{4}{*}{10\%}
& LLM-Pruner
& 8.3402 & 31.7846
& 0.7037 & 0.7693 & 0.5526 & 0.6780
& \textbf{0.7306} & \textbf{0.4249} & 0.3000
& 0.5942 \\
& FLAP
& 19.0166 & 64.8369
& 0.6630 & 0.7595 & 0.5594 & 0.6725
& 0.6570 & 0.3788 & 0.2300
& 0.5600 \\
& LoRAP
& 7.5938 & 27.6150
& 0.7621 & 0.7655 & \textbf{0.5685} & 0.6819
& 0.7155 & 0.4070 & 0.3120
& 0.6018 \\
& CASP
& \textbf{7.3818} & \textbf{25.5896}
& \textbf{0.7820} & \textbf{0.7699} & 0.5671 & \textbf{0.6827}
& 0.7247 & 0.4130 & \textbf{0.3200}
& \textbf{0.6085} \\
\midrule

\multirow{4}{*}{20\%}
& LLM-Pruner
& 12.5446 & 51.5916
& 0.5633 & 0.7476 & 0.5194 & 0.6606
& 0.7033 & \textbf{0.4087} & 0.3060
& 0.5584 \\
& FLAP
& 22.0596 & 74.0460
& 0.5878 & 0.7280 & 0.5023 & 0.6440
& 0.6376 & 0.3387 & 0.3120
& 0.5037 \\
& LoRAP
& 8.5047 & 35.7364
& 0.6850 & 0.7519 & \textbf{0.5485} & \textbf{0.6756}
& 0.7002 & 0.3925 & 0.3160
& 0.5814 \\
& CASP
& \textbf{8.1311} & \textbf{28.4360}
& \textbf{0.7440} & \textbf{0.7677} & 0.5345 & 0.6646
& \textbf{0.7176} & 0.3942 & \textbf{0.3300}
& \textbf{0.5932} \\
\midrule

\multirow{4}{*}{30\%}
& LLM-Pruner
& 17.7254 & 76.9957
& 0.4596 & 0.7106 & 0.4632 & 0.5927
& 0.6604 & 0.3635 & 0.2780
& 0.5040 \\
& FLAP
& 26.6610 & 87.2475
& 0.5410 & 0.6926 & 0.4480 & 0.6330
& 0.6124 & 0.3268 & 0.2640
& 0.5025 \\
& LoRAP
& 10.2787 & 55.8829
& \textbf{0.6731} & \textbf{0.7405} & \textbf{0.5199} & \textbf{0.6448}
& 0.6498 & 0.3515 & \textbf{0.2980}
& \textbf{0.5539} \\
& CASP
& \textbf{9.7317} & \textbf{33.1801}
& 0.6639 & 0.7367 & 0.4924 & 0.6069
& \textbf{0.6780} & \textbf{0.3771} & 0.2900
& 0.5493 \\
\midrule

\multirow{4}{*}{50\%}
& LLM-Pruner
& 91.7925 & 460.7304
& 0.4896 & 0.6115 & 0.3272 & 0.5320
& 0.4390 & \textbf{0.2654} & 0.1780
& 0.4061 \\
& FLAP
& 53.4376 & 206.0512
& 0.3838 & 0.6148 & 0.3366 & 0.5264
& 0.4327 & 0.2560 & 0.1820
& 0.3903 \\
& LoRAP
& 22.9549 & 129.2947
& \textbf{0.6058} & 0.6235 & 0.3591 & \textbf{0.5556}
& 0.4141 & 0.2509 & 0.1960
& 0.4293 \\
& CASP
& \textbf{18.1812} & \textbf{67.9484}
& 0.6034 & \textbf{0.6572} & \textbf{0.3762} & 0.5541
& \textbf{0.4596} & 0.2577 & \textbf{0.2360}
& \textbf{0.4492} \\
\bottomrule
\end{tabular}%
}
\label{tab:main_large_table_sheet_llama7b}
\end{table*}

\begin{table*}[ht]
\centering
\scriptsize
\setlength{\tabcolsep}{3.2pt}
\renewcommand{\arraystretch}{1.15}
\caption{Zero-shot performance of the compressed LLaMA2-13B. The average score is computed across seven datasets. The "\textbf{bolded}" represents the best result under the same pruning ratio.}
\resizebox{\textwidth}{!}{%
\begin{tabular}{cc|cc|ccccccccc|c}
\toprule
Ratio & Method
& WikiText2 $\downarrow$ & PTB $\downarrow$
& BoolQ $\uparrow$ & PIQA $\uparrow$ & HellaSwag $\uparrow$ & WinoGrande $\uparrow$
& ARC-e $\uparrow$ & ARC-c $\uparrow$ & OBQA $\uparrow$
& Average $\uparrow$ \\
\midrule

0\% & Ground Truth
& 4.8838 & 28.9402
& 0.8214 & 0.7943 & 0.6019 & 0.7238
& 0.7896 & 0.4735 & 0.3460
& 0.6501 \\
\midrule

\multirow{4}{*}{10\%}
& LLM-Pruner
& 5.6887 & 34.9768
& 0.7333 & 0.7845 & 0.5838 & 0.7088
& 0.7748 & 0.4599 & \textbf{0.3460}
& 0.6273 \\
& FLAP
& 12.1349 & 59.7305
& 0.7673 & 0.7802 & 0.5688 & 0.7072
& 0.7386 & 0.4215 & 0.3420
& 0.6179 \\
& LoRAP
& 5.9488 & 32.8577
& \textbf{0.8208} & 0.7878 & \textbf{0.6000} & 0.7206
& 0.7786 & 0.4590 & \textbf{0.3460}
& 0.6447 \\
& CASP
& \textbf{5.1634} & \textbf{32.7983}
& 0.8196 & \textbf{0.7927} & 0.5971 & \textbf{0.7293}
& \textbf{0.7824} & \textbf{0.4710} & 0.3420
& \textbf{0.6477} \\
\midrule

\multirow{4}{*}{20\%}
& LLM-Pruner
& 6.6494 & 45.8865
& 0.6514 & \textbf{0.7884} & 0.5619 & 0.7009
& \textbf{0.7580} & 0.4428 & 0.3320
& 0.6051 \\
& FLAP
& 14.1319 & 68.7494
& 0.6826 & 0.7546 & 0.5182 & 0.6788
& 0.6797 & 0.3660 & 0.3120
& 0.5703 \\
& LoRAP
& 6.4953 & 37.9670
& \textbf{0.8128} & 0.7786 & \textbf{0.5885} & 0.7080
& 0.7431 & \textbf{0.4548} & \textbf{0.3540}
& \textbf{0.6343} \\
& CASP
& \textbf{5.6488} & \textbf{35.4583}
& 0.8083 & 0.7737 & 0.5821 & \textbf{0.7277}
& 0.7391 & 0.4300 & 0.3440
& 0.6293 \\
\midrule

\multirow{4}{*}{30\%}
& LLM-Pruner
& 9.5808 & 71.2094
& 0.6370 & \textbf{0.7617} & 0.5172 & 0.6377
& \textbf{0.7214} & 0.3925 & \textbf{0.3140}
& 0.5688 \\
& FLAP
& 16.6514 & 81.3237
& 0.6434 & 0.7182 & 0.4711 & 0.6417
& 0.6275 & 0.3549 & 0.2780
& 0.5335 \\
& LoRAP
& 7.4907 & 52.6091
& 0.7196 & 0.7590 & \textbf{0.5584} & \textbf{0.6961}
& 0.6665 & \textbf{0.4181} & 0.3000
& 0.5882 \\
& CASP
& \textbf{6.3946} & \textbf{39.0195}
& \textbf{0.7284} & 0.7557 & 0.5417 & 0.6772
& 0.7159 & 0.4147 & 0.3120
& \textbf{0.5922} \\
\midrule

\multirow{4}{*}{50\%}
& LLM-Pruner
& 25.3904 & 191.3121
& 0.4627 & 0.6742 & 0.3920 & 0.5517
& 0.5135 & 0.2730 & 0.2420
& 0.4442 \\
& FLAP
& 27.8860 & 156.7555
& 0.5810 & 0.6583 & 0.3830 & 0.5864
& 0.4912 & \textbf{0.3055} & 0.2540
& 0.4656 \\
& LoRAP
& 15.8889 & 247.5762
& 0.6690 & 0.6757 & 0.4171 & 0.5948
& \textbf{0.5800} & 0.2863 & 0.2320
& 0.4936 \\
& CASP
& \textbf{9.8272} & \textbf{54.0625}
& \textbf{0.6826} & \textbf{0.6834} & \textbf{0.4273} & \textbf{0.6361}
& 0.5417 & 0.2952 & \textbf{0.2760}
& \textbf{0.5060} \\
\bottomrule
\end{tabular}%
}
\label{tab:main_large_table_sheet_llama13b}
\end{table*}

\section{Experiments}
\label{sec:exp}
\subsection{Experimental Settings}
\label{sec:exp_setup}

\paragraph{Models and Evaluation.}
We evaluate our method on representative open-source large language models, including the LLaMA2 family (7B and 13B)~\cite{touvron2023llama2}, Vicuna-7B~\cite{chiang2023vicuna}, and the recent GQA-based LLaMA3-8B model~\cite{grattafiori2024llama}. To comprehensively assess the performance of the compressed models, we employ two standard metrics. First, we report the zero-shot perplexity (PPL) on the test sets of WikiText2~\cite{merity2017pointer} and PTB~\cite{marcus1993building} datasets to measure the language modeling capability. Second, following previous structured pruning works~\cite{ma2023llm,li2024lorap}, we evaluate the zero-shot accuracy on seven common sense reasoning datasets, including BoolQ~\cite{clark2019boolq}, PIQA~\cite{bisk2020piqa}, HellaSwag~\cite{zellers2019hellaswag}, WinoGrande~\cite{sakaguchi2021winogrande}, ARC-easy~\cite{clark2018think}, ARC-challenge~\cite{clark2018think}, and OpenbookQA~\cite{mihaylov2018can}.

\paragraph{Calibration data and implementation details.}
To estimate the activation statistics, we randomly sample 128 sequences from the WikiText2~\cite{merity2017pointer} training set, each with a context length of 2048 tokens. Similarly, we employ a sequence length of 2048 tokens when measuring perplexity on the test datasets. We evaluate our method across four distinct sparsity levels: 10\%, 20\%, 30\%, and 50\%. For the layer-wise sparsity allocation, we adjust the temperature parameter $\alpha$ based on the global budget: we employ a smaller $\alpha$ for low compression regimes to yield a smoother sparsity distribution, while adopting a larger $\alpha$ for high compression regimes to aggressively allocate higher sparsity to less sensitive layers.

\paragraph{Baselines.}
We compare our proposed method with the following representative structured pruning baselines:
\begin{itemize}
    \item \textbf{LLM-Pruner}~\cite{ma2023llm} is the pioneering structured pruning method for LLMs. It utilizes first-order Taylor expansion to estimate the importance of coupled structures and removes the least important groups based on dependency detection.
    \item \textbf{LoRAP}~\cite{li2024lorap} proposes a differentiated compression strategy. It applies low-rank approximation to the Multi-Head Attention (MHA) layers and gradient-free structured pruning to the Feed-Forward Networks (FFN) based on activation-weighted importance.
    \item \textbf{FLAP}~\cite{an2024fluctuation} introduces a fluctuation-based pruning metric. It evaluates feature importance by measuring the insensitivity of the model to the removal of specific channels, focusing on preserving features with high variance.
\end{itemize}

\begin{remark}
To rigorously assess the intrinsic quality of the pruning masks identified by CASP, we primarily focus our evaluation on the post-training regime. For completeness, we provide supplementary results demonstrating the performance recovery after fine-tuning in Appendix~\ref{appendix:finetune}, confirming that our method also serves as a robust initialization for further optimization.
\end{remark}

\subsection{The Off-Diagonal Energy Ratio of
LLM Layers}
\label{subsection:off_diag}

{As discussed previously}, the independence assumption—often adopted by existing pruning methods to score units in isolation—is insufficient for LLMs. To empirically validate this hypothesis, we quantify the magnitude of cross-unit dependencies within the reconstruction error.

We define the off-diagonal energy ratio $\rho$  as the proportion of the total error energy matrix $Q$ constituted by interaction terms:
\begin{equation}
    \rho = \frac{\sum_{i \neq j} |Q_{ij}|}{\sum_{i,j} |Q_{ij}|},
\end{equation}
where $Q_{ij}$ represents the entries of the interaction matrix {defined} in Section \ref{sec:modeling}. A low $\rho$ would suggest that units act independently (validating the diagonal assumption), while a high $\rho$ indicates that the error landscape is dominated by cross-channel correlations.

\begin{figure}[t]
    \centering
    \includegraphics[width=1\linewidth]{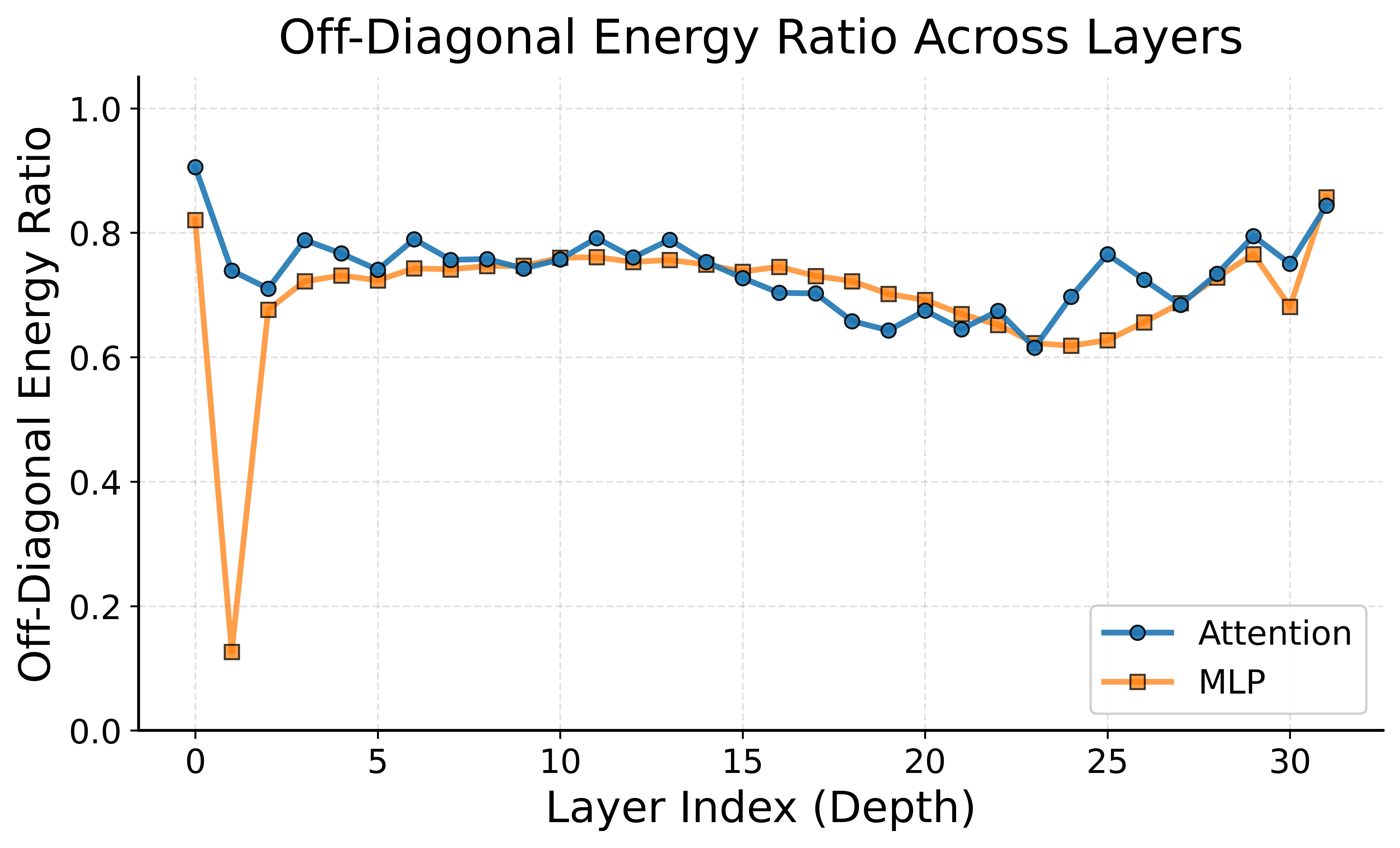}
    \caption{\textbf{Off-diagonal energy ratio across layers in LLaMA2-7B.} We report the proportion of interaction (off-diagonal) terms in the reconstruction energy for both attention-head pruning and FFN-channel pruning. The consistently high ratio demonstrates that cross-unit interactions dominate the pruning error in most layers, invalidating the independence assumption.}
    \label{fig:offdiag_ratio}
\end{figure}

Figure~\ref{fig:offdiag_ratio} illustrates the layer-wise evolution of $\rho$ on LLaMA2-7B. We observe that $\rho$ consistently exceeds 0.6 across the vast majority of layers in both attention and FFN blocks, implying that off-diagonal terms dominate the total error energy. These {empirical results} provide compelling evidence that approximating $Q$ as a diagonal matrix neglects critical structural information. Consequently, our proposed CASP method is essential to explicitly model these cross-unit dependencies.

\subsection{Zero-Shot Performance}
\label{sec:exp_main}

Table~\ref{tab:main_large_table_sheet_llama7b} and Table~\ref{tab:main_large_table_sheet_llama13b} summarize the zero-shot performance results for LLaMA2-7B and LLaMA2-13B, respectively (results for other models are provided in Appendix~\ref{appendix:exp_vicuna} and~\ref{appendix:gqa}). We compare CASP against representative baselines introduced above across varying pruning ratios. As demonstrated in the tables, CASP outperforms baseline algorithms in the vast majority of cases in terms of both PPL metrics and average accuracy on common sense reasoning datasets.

In low compression regimes (10\% and 20\%), CASP exhibits strong performance preservation, indicating its ability to effectively identify redundant structures. Notably, at a 20\% pruning ratio, CASP incurs only marginal accuracy loss compared to the unpruned model while maintaining lower PPL scores than baselines. As the compression ratio increases to the aggressive level of 50\%, the advantage of CASP becomes more evident. 
While baseline methods suffer from catastrophic performance degradation—manifested by exploding PPL values—CASP successfully retains competitive functionality and linguistic coherence. For instance, on LLaMA2-13B with a 50\% pruning ratio, the WikiText2 PPL of CASP remains below 10 (specifically 9.83).
\subsection{Ablation Study}
To systematically evaluate the contribution of each component in our proposed CASP framework, we conduct an ablation study on LLaMA2-7B (Table~\ref{tab:ablation_study}), comparing the full method against two variants: \textbf{w/o Greedy}, which replaces our interaction-aware greedy selection with an independent strategy that treats the interaction matrix   $ Q $   as diagonal, and \textbf{w/o Allocation}, which applies a uniform pruning ratio instead of adaptive layer-wise allocation.

\paragraph{Greedy Interaction Algorithm.}
As shown in Table~\ref{tab:ablation_study}, the greedy interaction algorithm exhibits a moderate advantage over the independent selection strategy at 30\% sparsity, and this gap becomes markedly wider at 50\% sparsity. Specifically, CASP achieves an average zero-shot accuracy that is more than 4\% higher than the \textbf{w/o Greedy} variant, with notably lower perplexity on both WikiText2 and PTB. This confirms that the independence assumption used in prior work is inadequate for LLMs; by accounting for cross-unit dependencies, CASP makes significantly more robust pruning decisions.

\paragraph{Layer-Wise Allocation.}
As shown in Table~\ref{tab:ablation_study}, applying a uniform sparsity ratio severely degrades pruning performance, especially under high pruning ratios. The \textbf{w/o Allocation} variant catastrophically fails at 50\% sparsity: WikiText2 PPL surges to 58.81, and average zero-shot accuracy collapses to 33.99\%. This underscores the importance of leveraging layer-wise sensitivity to allocate sparsity adaptively.

\begin{table}[t]
\centering
\fontsize{10}{12}\selectfont
% 如果表格太宽，可以取消注释这一行
\setlength{\tabcolsep}{3pt} % 调整列间距
\caption{
Ablation study of different components in our proposed method (CASP) at 30\% and 50\% pruning ratios. 
 The "\textbf{bolded}" represents the best result under the same pruning ratio.
}
\label{tab:ablation_study}
\begin{tabular}{c|l|cc|c}
\toprule
\textbf{Ratio} & \textbf{Method} & \textbf{WikiText2}$\downarrow$ & \textbf{PTB}$\downarrow$ & \textbf{Avg.} \\
\midrule
\multirow{3}{*}{30\%} 
& w/o Greedy & 10.40 & \textbf{31.11}& 0.5211 \\
& w/o Allocation & 11.45 & 32.47 & 0.4114 \\
& CASP     & \textbf{9.73} & 33.18 & \textbf{0.5493} \\
\midrule
\multirow{3}{*}{50\%} 
& w/o Greedy & 22.94 & 72.63 & 0.4088 \\
& w/o Allocation & 58.81 & 159.88 & 0.3399 \\
& CASP & \textbf{18.18} & \textbf{67.95} & \textbf{0.4492} \\
\bottomrule
\end{tabular}
\vspace{-8pt}
\end{table}
\section{Conclusion}
In this paper, we propose Correlation-Aware Structured Pruning (CASP), a post-training approach designed to address the limitations of the independence assumption commonly adopted in structured pruning. By formulating the pruning objective as a cardinality-constrained binary quadratic program, we explicitly model the cross-unit dependencies inherent in LLMs. %To solve this efficiently, we developed a greedy interaction algorithm. Furthermore, to enhance end-to-end performance, we incorporated a gradient-based layer-wise allocation strategy that adaptively distributes sparsity based on sensitivity. 
{We further develop a greedy interaction algorithm for efficient optimization, together with a gradient-based layer-wise strategy for sensitivity-aware sparsity assignment.}
{Extensive experiments on mainstream LLMs demonstrate that CASP achieves robust performance preservation and favorable accuracy-efficiency trade-offs, particularly under high sparsity ratios.}
%Extensive experiments on mainstream LLMs demonstrate that CASP achieves superior accuracy-efficiency trade-offs compared to representative baselines.

%{Beyond aggregate metrics such as perplexity and zero-shot accuracy, investigating how structured pruning affects specific capabilities (e.g., factual recall, multi-step reasoning, and safety-related behaviors) could be considered as a potential extension of this work.}

\section{Limitations}

{Due to scope constraints, the experiments in this work are conducted on dense decoder-only Transformer LLMs, which remain the dominant architecture for modern large language models. The reported results therefore primarily validate the effectiveness of the proposed method within this architectural setting.} Extending the proposed approach to newer or structurally different paradigms, such as mixture-of-experts (MoE) models or multimodal models, remains an important direction for future work.

In addition, the current evaluation primarily relies on perplexity and zero-shot accuracy benchmarks. While these metrics are widely adopted for assessing compressed language models, they may not fully capture the impact of structured pruning on more complex capabilities, such as multi-step reasoning, long-context understanding, factual robustness, or safety-related behaviors. 

\section{Ethical Considerations}

This work studies structured pruning for improving the efficiency of LLMs. Our experiments are conducted using publicly available pretrained models and benchmark datasets, without collecting or involving personal or sensitive user data. By reducing the computational and deployment cost of LLMs, efficient compression methods may help lower energy consumption and improve the accessibility of LLMs in resource-constrained environments.
\bibliography{ref}

%%%%%%%%%%%%%%%%%%%%%%%%%%%%%%%%%%%%%%%%%%%%%%%%%%%%%%%%%%%%%%%%%%%%%%%%%%%%%%%
%%%%%%%%%%%%%%%%%%%%%%%%%%%%%%%%%%%%%%%%%%%%%%%%%%%%%%%%%%%%%%%%%%%%%%%%%%%%%%%
% APPENDIX
%%%%%%%%%%%%%%%%%%%%%%%%%%%%%%%%%%%%%%%%%%%%%%%%%%%%%%%%%%%%%%%%%%%%%%%%%%%%%%%
%%%%%%%%%%%%%%%%%%%%%%%%%%%%%%%%%%%%%%%%%%%%%%%%%%%%%%%%%%%%%%%%%%%%%%%%%%%%%%%

\newpage

\appendix
 \begin{figure*}[ht]
    \centering
    \includegraphics[width=0.6\linewidth]{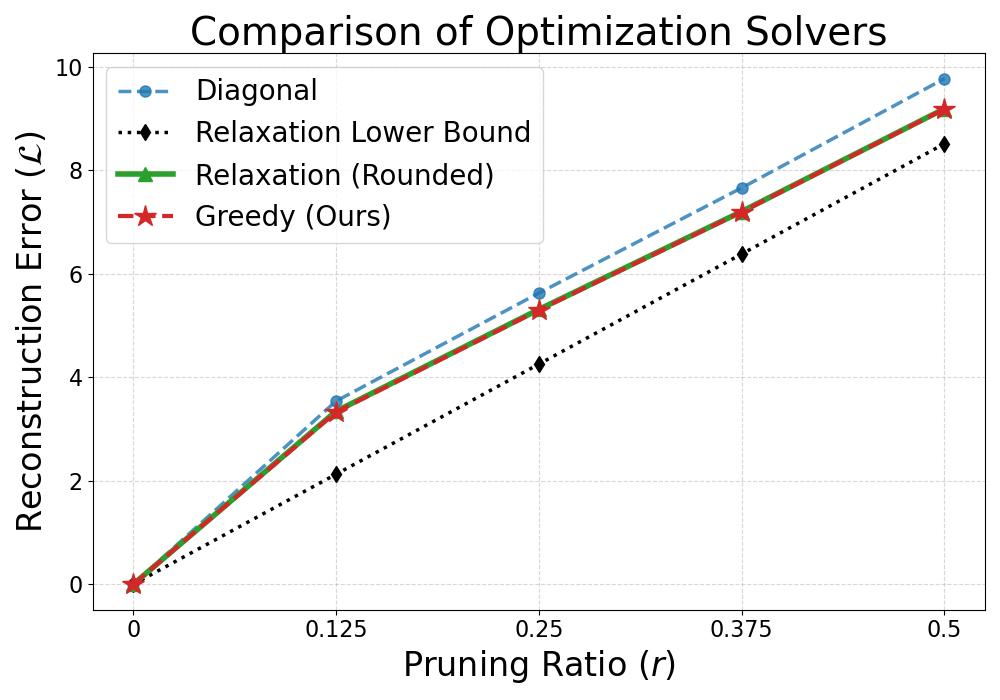} % Adjusted width for better visibility
    \caption{\textbf{Comparison of Optimization Methods.} Reconstruction error ($\mathcal{L}$) on the first FFN layer of LLaMA2-7B under varying pruning ratios ($r$). We plot the \textit{Diagonal} baseline (blue), the theoretical \textit{Relaxation Lower Bound} (black dotted), the projected \textit{Relaxation} solution (green solid), and our \textit{Greedy} interaction algorithm (red dashed).}
    \label{fig:relax_bound}
\end{figure*}
\section{Theoretical Analysis and Derivations}
\subsection{The Continuous Relaxation Method}
\label{appendix:continuous_relaxation}

We provide a supplementary introduction to the continuous relaxation method for the optimization problem formulated in ~\eqref{eq:qubo_card} Specifically, we relax the binary constraint $z \in \{0, 1\}^{d_m}$ to the continuous interval $z \in [0, 1]^{d_m}$, yielding the following optimization problem:
\begin{align}
\label{eq:relaxed_qp}
\min_{z\in [0,1]^{d_m}} \quad & z^\top Q z \\
\text{s.t.}\quad & \mathbf{1}^\top z = k. \nonumber
\end{align}
Since the reconstruction error matrix $Q$ is positive semi-definite by construction, the problem in Eq. \eqref{eq:relaxed_qp} is a standard convex Quadratic Programming (QP) problem, which can theoretically be solved in polynomial time \cite{boyd2004convex, kozlov1979polynomial}. Upon obtaining the optimal continuous solution $z^\star_{\text{relax}}$, we can derive a feasible binary solution for the original problem by projecting $z^\star_{\text{relax}}$ onto the discrete constraint set (e.g., by retaining the indices corresponding to the $k$ largest values in $z^\star_{\text{relax}}$).

However, in the context of Large Language Models, the dimensionality of the interaction matrix $Q$ presents a significant computational challenge. For instance, in the FFN layers of LLaMA2-7B, the intermediate dimension $d_m$ is 11,008, which implies that $Q$ is a square matrix of order 11,008. Solving such a large-scale Quadratic Programming problem for every single layer incurs a substantial computational cost. Therefore, in practical implementation, we adopt the more efficient Greedy Interaction Search. Nevertheless, investigating this relaxation holds theoretical value as it provides a rigorous baseline for assessing the quality of discrete algorithms.

Let $\mathcal{L}^{\star}_{\text{relax}}$ denote the optimal objective value of the continuous problem \eqref{eq:relaxed_qp}, and let $\mathcal{L}^{\star}_{\text{discrete}}$ be the global minimum of the original binary problem \eqref{eq:qubo_card}. Since the discrete feasible set is strictly contained within the continuous domain (i.e., $\{0,1\}^{d_m} \subset [0,1]^{d_m}$), the relationship $f_{\text{relax}} \le f_{\text{discrete}}$ always holds. Consequently, $\mathcal{L}^{\star}_{\text{relax}}$ constitutes a lower bound for any valid binary solution, including the solution obtained by our proposed Greedy Interaction Search and the solution obtained by projecting the continuous relaxation. This leads to the inequality chain:
\begin{equation*}
    \mathcal{L}^{\star}_{\text{relax}} \le \mathcal{L}^{\star}_{\text{discrete}} \le \mathcal{L}_{\text{greedy}}, \mathcal{L}_{\text{relax}},
\end{equation*}
where $\mathcal{L}_{\text{greedy}}$ and $\mathcal{L}_{\text{relax}}$ represent the objective values evaluated using the feasible binary solutions derived from the Greedy Interaction Search and the projection of the continuous optimal solution, respectively.

Figure \ref{fig:relax_bound} illustrates the reconstruction error trends on the first FFN layer of LLaMA2-7B. We observe that our efficient Greedy Interaction Search yields error rates nearly identical to the projected solution from the computationally expensive continuous relaxation, with both significantly outperforming the diagonal baseline. Furthermore, the gap between our greedy solution and the theoretical lower bound remains narrow across all sparsity levels.

\subsection{Efficient Implementation and Complexity}
\label{appendix:complexity}

Directly evaluating the marginal gain $\Delta_i(S)$ for all candidates at each step using \eqref{eq:marginal_cost} is computationally expensive. In a naive implementation, calculating the interaction term $\sum_{j \in S} Q_{ij}$ for all $d_m - |S|$ candidates requires $O(|S| \cdot d_m)$ operations. Over $k$ pruning steps, this accumulates to a total complexity of $O(\sum_{t=1}^k t \cdot d_m) = O(k^2 d_m)$, which is prohibitive for large layers.

Here, we derive the efficient update rule used in Algorithm \ref{alg:greedy_interaction} and prove its equivalence to the naive greedy selection. We maintain an auxiliary vector $u \in \mathbb{R}^{d_m}$ initialized to zero, where $u_i^{(t)} \triangleq \sum_{j \in S_t} Q_{ij}$ represents the accumulated interaction cost at step $t$. When a new unit $i^*$ is selected and added to the set (i.e., $S_{t+1} = S_t \cup \{i^*\}$), the interaction term for any remaining candidate $i$ updates linearly:
\begin{equation}
    \sum_{j \in S_{t+1}} Q_{ij} = \sum_{j \in S_t} Q_{ij} + Q_{i, i^*}.
\end{equation}
This allows us to update the auxiliary vector via a simple vector addition: $u^{(t+1)} \leftarrow u^{(t)} + Q_{:, i^*}$, where $Q_{:, i^*}$ is the column of $Q$ corresponding to the selected unit. Consequently, the marginal cost evaluation $\Delta_i = Q_{ii} + 2u_i$ becomes an $O(1)$ operation, and the update step takes $O(d_m)$. This efficient implementation reduces the total complexity from quadratic $O(k^2 d_m)$ to linear $O(k \cdot d_m)$ with respect to the pruning budget $k$, ensuring scalability.
% (Continue with the derivation of the update step u <- u + Q[:, i*])

\subsection{Justification for Block-wise Pruning}
\label{appendix:block_wise}

For Multi-Head Attention (MHA), pruning is typically performed at the granularity of attention heads rather than individual neurons to preserve the structural integrity of the attention mechanism. Let $\mathcal{B}_h$ be the set of channel indices corresponding to the $h$-th attention head, where $h \in \{1, \dots, H\}$. The decision to prune head $h$ is represented by a binary variable $s_h \in \{0, 1\}$.

The relationship between the channel-wise mask $z$ and the head-wise mask $s$ is given by the constraint: $z_i = s_h$ for all $i \in \mathcal{B}_h$. Substituting this into the general reconstruction error objective $\mathcal{L}(S) = \sum_{i, j} z_i z_j Q_{ij}$, we derive the block-wise objective:

\begin{equation}
\begin{aligned}
    \mathcal{L}_{\text{head}}(s) &= \sum_{a=1}^{H} \sum_{b=1}^{H} \sum_{i \in \mathcal{B}_a} \sum_{j \in \mathcal{B}_b} z_i z_j Q_{ij} \\
    &= \sum_{a=1}^{H} \sum_{b=1}^{H} s_a s_b \left( \sum_{i \in \mathcal{B}_a} \sum_{j \in \mathcal{B}_b} Q_{ij} \right).
\end{aligned}
\end{equation}

By defining the block interaction matrix $\tilde{Q} \in \mathbb{R}^{H \times H}$ with entries $\tilde{Q}_{ab} \triangleq \sum_{i \in \mathcal{B}_a} \sum_{j \in \mathcal{B}_b} Q_{ij}$, the objective simplifies to the quadratic form $s^\top \tilde{Q} s$.
This derivation proves that minimizing the reconstruction error under the block-wise constraint is mathematically equivalent to applying our Greedy Interaction Search on the aggregated matrix $\tilde{Q}$. Consequently, our method naturally extends to attention head pruning without any heuristic approximations.

\section{Supplementary Experiments}
\subsection{Layer-Wise Gradient Sensitivity}
\label{appedix:sensitivity}

\begin{figure*}[ht]
    \centering
    \includegraphics[width=0.9\linewidth]{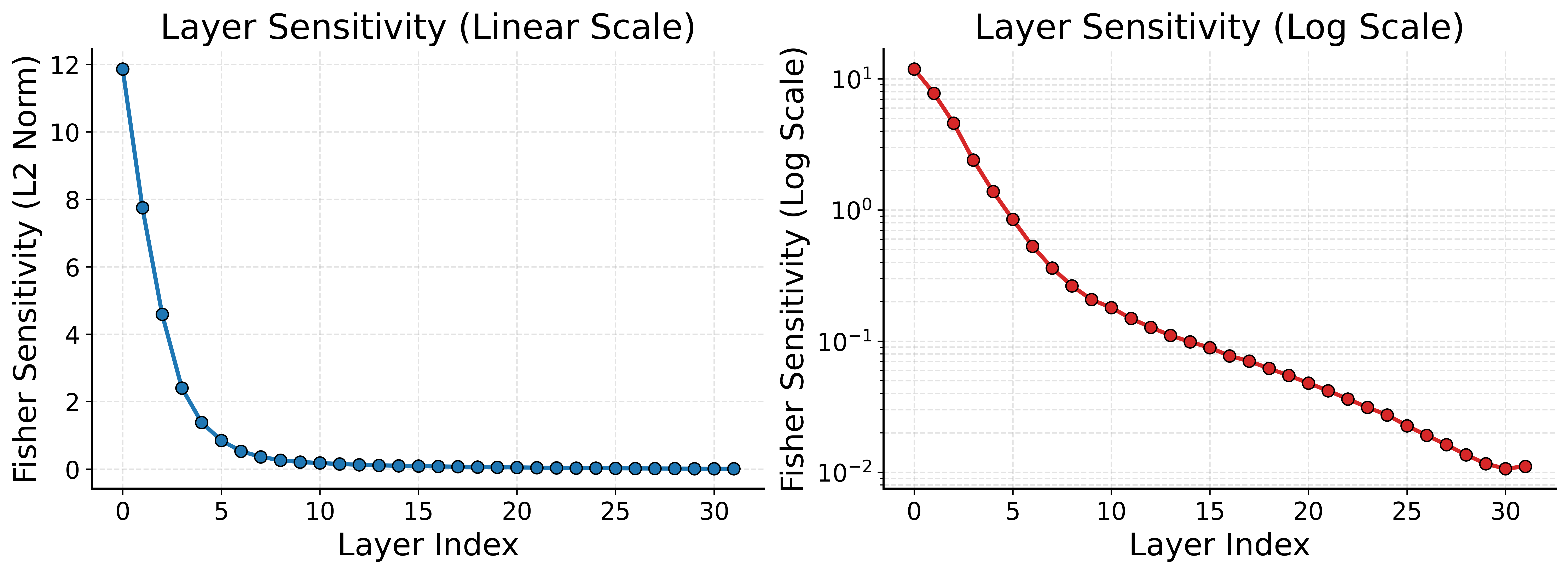}
    \caption{Layer-wise gradient sensitivity in LLaMA2-7B. We compute each layer’s sensitivity as the squared $\ell_2$ norm of the gradient of the final objective with respect to the layer output, and report it across depth (linear and log scales). The sensitivity generally decreases with depth, motivating an increasing pruning ratio for deeper layers.}
    \label{fig:layer_sensitivity}
\end{figure*}

We analyze the layer-wise sensitivity of LLaMA2-7B to justify our non-uniform sparsity allocation. Figure~\ref{fig:layer_sensitivity} reports the squared $\ell_2$ norm of the gradients with respect to each layer's output. We observe a sharp decline in sensitivity from the shallow layers to the deep layers, spanning several orders of magnitude as shown in the log-scale plot. This confirms that early layers are critical for feature extraction and highly sensitive to perturbations, whereas deeper layers possess greater redundancy. Accordingly, our designed compression ratio allocation exhibits a layer-wise increasing trend.

\subsection{Pruning Time}
\label{appendix:time}

To quantify the computational overhead introduced by the pruning algorithm, we report the wall-clock pruning time on LLaMA2-7B under different pruning ratios. We use the same calibration setting as in the main experiments, i.e., $128$ sequences with sequence length $2048$. Specifically, we test pruning ratios of $0.1$, $0.2$, $0.3$, and $0.5$. For each setting, we repeat the pruning process $10$ times and report the averaged runtime. As shown in Table~\ref{tab:pruning_time}, the pruning time is nearly insensitive to the pruning ratio. This suggests that the main runtime bottleneck lies in the calibration forward passes and interaction matrix construction, rather than in the greedy pruning algorithm itself; nevertheless, the entire pruning procedure can be completed within a few minutes, demonstrating its practical efficiency.
\begin{table}[ht]
\centering
\caption{Average pruning time of our method on LLaMA2-7B under different pruning ratios. Each result is averaged over $10$ runs.}
\label{tab:pruning_time}
\begin{tabular}{c|c}
\toprule
Ratio & Time (s) \\
\midrule
$10\%$ & 184.45 \\
$20\%$ & 185.33 \\
$30\%$ & 185.35 \\
$50\%$ & 186.59 \\
\bottomrule
\end{tabular}
\end{table}

\subsection{Inference Efficiency Analysis}
\label{appendix:efficiency}

To validate the practical efficiency gains of our method, we evaluate the inference latency and computational cost of LLaMA2-7B pruned via CASP. We focus on three key metrics: model parameter counts, theoretical computational complexity measured in Multiply-Accumulate operations (MACs), and actual wall-clock inference latency. Specifically, the theoretical MACs are calculated based on processing a single batch with a sequence length of 128 during the prefilling phase. For the wall-clock latency measurements, we simulate a realistic inference scenario: the prefilling latency is measured using a batch size of 32 with an input sequence length of 128, and the decoding latency is recorded while generating 128 subsequent tokens. All latency measurements are averaged over 10 runs.

Table~\ref{tab:latency} presents the detailed performance data for LLaMA2-7B at sparsity levels of 20\%  and 50\% . We observe that CASP achieves substantial reductions in computational overhead that translate effectively into real-world speedups. These results confirm that the structured nature of CASP facilitates direct acceleration on standard hardware, effectively bridging the gap between theoretical complexity reduction and practical inference speed.

\begin{table*}[ht]
\centering
\caption{Inference latency and parameter counts for LLaMA2-7B at different sparsity levels.}
\label{tab:latency}
\begin{tabular}{lccccc}
\toprule
Ratio & Method & Params (B) & MACs(G) & Prefilling(s) & Decoding(s) \\
\midrule
0\% & Dense  & 6.74 & 845.71 & 0.321 & 2.694 \\
20\%  & CASP   & 5.44 & 624.80 & 0.260 & 2.243 \\
50\%  & CASP   & 3.51 & 397.62 & 0.175 & 1.870 \\
\bottomrule
\end{tabular}
\end{table*}

\subsection{Performance Recovery with Fine-tuning}
\label{appendix:finetune}

Here, we provide supplementary experiments to evaluate the performance recovery of pruned models via fine-tuning. Specifically, we apply LoRA fine-tuning to LLaMA2-7B models pruned at 50\% sparsity using both CASP and LoRAP. The models are fine-tuned on the cleaned version of the Alpaca dataset~\cite{taori2023stanford} for 2 epochs, employing a standard learning rate of $1\times 10^{-4}$ and a batch size of 128. The results, summarized in Table~\ref{tab:finetune}, demonstrate that both methods exhibit significant performance improvements after fine-tuning. Notably, our method outperforms LoRAP following this recovery phase.

\begin{table*}[ht]
\centering
\scriptsize
\setlength{\tabcolsep}{3.2pt}
\renewcommand{\arraystretch}{1.15}
\caption{Performance recovery via fine-tuning on LLaMA2-7B at 50\% sparsity. The average score is computed across seven datasets. The "\textbf{bolded}" represents the best result under the same pruning ratio.}
\resizebox{\textwidth}{!}{%
\begin{tabular}{cc|cc|ccccccccc|c}
\toprule
Ratio & Method
& WikiText2 $\downarrow$ & PTB $\downarrow$
& BoolQ $\uparrow$ & PIQA $\uparrow$ & HellaSwag $\uparrow$ & WinoGrande $\uparrow$
& ARC-e $\uparrow$ & ARC-c $\uparrow$ & OBQA $\uparrow$
& Average $\uparrow$ \\
\midrule
\multirow{2}{*}{\shortstack{50\% \\ w/o tune}} 
& LoRAP
& 22.9549 & 129.2947
& \textbf{0.6058} & 0.6235 & 0.3591 & \textbf{0.5556}
& 0.4141 & 0.2509 & 0.1960
& 0.4293 \\
& CASP
& \textbf{18.1812} & \textbf{67.9484}
& 0.6034 & \textbf{0.6572} & \textbf{0.3762} & 0.5541
& \textbf{0.4596} & \textbf{0.2577} & \textbf{0.2360}
& \textbf{0.4492} \\
\midrule
\multirow{2}{*}{\shortstack{50\% \\ w/ tune}} 
& LoRAP
& 12.9873 & 59.5412
& 0.6181 & \textbf{0.6975} & 0.4226 & \textbf{0.5951}
& 0.5610 & \textbf{0.2841} & 0.2460
& 0.4892 \\
& CASP
& \textbf{12.5192} & \textbf{42.4517}
& \textbf{0.6386} & 0.6757 & \textbf{0.4349} & 0.5635
& \textbf{0.6232} & 0.2711 & \textbf{0.2480}
& \textbf{0.4936} \\
\bottomrule
\end{tabular}}
\label{tab:finetune}
\end{table*}

\subsection{Ablation Study of Temperature Parameter}
\begin{table}[ht]
\centering
\caption{
Ablation study of temperature parameter $\alpha$ in our proposed method (CASP) at  50\% pruning ratios. 
 The "\textbf{bolded}" represents the best result under the same pruning ratio.
}
\begin{tabular}{c|cc|c}
\toprule
\textbf{Value} & \textbf{WikiText2} & \textbf{PTB} & \textbf{Avg.} \\
\midrule
$\alpha=0.5 $   & \textbf{17.45} & 76.70 & 0.4074 \\
$\alpha=1.0$ & 18.18 & \textbf{67.95} & \textbf{0.4492} \\
$\alpha=1.5$ & 24.51  & 80.38 & 0.4255 \\
$\alpha=2.0$ & 25.49 & 78.82 & 0.4384 \\
\bottomrule
\end{tabular}
\label{tab:ablation_alpha}
\end{table}

In our gradient-based layer-wise allocation strategy, the temperature parameter $\alpha$ plays a pivotal role in controlling the sparsity distribution. A larger $\alpha$ allocates higher compression rates to less sensitive layers, while a smaller $\alpha$
 yields a more uniform sparsity profile. To investigate the impact of this hyperparameter, we conduct a sensitivity analysis on LLaMA2-7B at a 50\% pruning ratio, varying $\alpha$
 from 0.5 to 2.0. The results, summarized in Table~\ref{tab:ablation_alpha}, indicate that $\alpha=1.0$ strikes the most effective balance between PPL metrics and reasoning accuracy, outperforming other configurations in overall stability.

\subsection{Results on Vicuna-7B}
\label{appendix:exp_vicuna}

To evaluate the generalization of our method, we apply CASP to Vicuna-7B. Table~\ref{tab:main_large_table_sheet_vicuna7b} summarizes the zero-shot performance comparison. 
The experimental results exhibit a trend consistent with the LLaMA2 families discussed in the main text. CASP consistently outperforms baselines in terms of PPL metrics while maintaining competitive reasoning accuracy across varying pruning ratios. This advantage is particularly evident in the high compression regime. At a 50\% pruning ratio, CASP achieves an average accuracy of 0.4718 and stabilizes the WikiText2 PPL at 18.04, showing remarkable robustness compared to LoRAP, which suffers from severe degradation (PPL 49.71).

\begin{table*}[ht]
\centering
\scriptsize
\setlength{\tabcolsep}{3.2pt}
\renewcommand{\arraystretch}{1.15}
\caption{Zero-shot performance of the compressed Vicuna-7B. The average score is computed across seven datasets. The "\textbf{bolded}" represents the best result under the same pruning ratio.}
\resizebox{\textwidth}{!}{%
\begin{tabular}{cc|cc|ccccccccc|c}
\toprule
Ratio & Method
& WikiText2 $\downarrow$ & PTB $\downarrow$
& BoolQ $\uparrow$ & PIQA $\uparrow$ & HellaSwag $\uparrow$ & WinoGrande $\uparrow$
& ARC-e $\uparrow$ & ARC-c $\uparrow$ & OBQA $\uparrow$
& Average $\uparrow$ \\
\midrule

0\% & Ground Truth
& 6.7805 & 26.7653
& 0.8162 & 0.7688 & 0.5657 & 0.7080
& 0.7597 & 0.4428 & 0.3500
& 0.6302 \\
\midrule

\multirow{4}{*}{10\%}
& LLM-Pruner
& 8.3728 & 30.8669
& 0.7095 & 0.7535 & 0.5444 & 0.6772
& 0.7487 & 0.4300 & \textbf{0.3400}
& 0.6005 \\
& FLAP
& 18.1103 & 67.1568
& 0.7306 & 0.7524 & 0.5457 & \textbf{0.6827}
& 0.7214 & 0.3993 & 0.3460
& 0.5969 \\
& LoRAP
& 7.3890 & 29.7425
& \textbf{0.7991} & 0.7633 & \textbf{0.5609} & 0.6788
& 0.7472 & 0.4352 & 0.3380
& \textbf{0.6175} \\
& CASP
& \textbf{7.2673} & \textbf{28.3805}
& 0.7936 & \textbf{0.7715} & 0.5589 & 0.6819
& \textbf{0.7500} & \textbf{0.4394} & 0.3260
& 0.6173 \\
\midrule

\multirow{4}{*}{20\%}
& LLM-Pruner
& 11.4220 & 38.4145
& 0.6453 & 0.7503 & 0.5136 & 0.6314
& 0.7218 & 0.3899 & 0.3020
& 0.5649 \\
& FLAP
& 21.0494 & 76.3965
& 0.6520 & 0.7252 & 0.4911 & 0.6535
& 0.6814 & 0.3626 & 0.2840
& 0.5499 \\
& LoRAP
& 8.1470 & 34.3003
& 0.7554 & 0.7524 & \textbf{0.5432} & \textbf{0.6764}
& \textbf{0.7344} & 0.4053 & \textbf{0.3160}
& \textbf{0.5976} \\
& CASP
& \textbf{8.0521} & \textbf{31.4143}
& \textbf{0.7697} & \textbf{0.7524} & 0.5300 & 0.6677
& 0.7214 & \textbf{0.4061} & 0.2980
& 0.5922 \\
\midrule

\multirow{4}{*}{30\%}
& LLM-Pruner
& 148.3910 & 52.7120
& 0.5505 & \textbf{0.7269} & 0.4714 & 0.5927
& \textbf{0.6717} & 0.3618 & 0.2800
& 0.5221 \\
& FLAP
& 25.2421 & 90.3694
& 0.6284 & 0.6937 & 0.4469 & 0.6188
& 0.6317 & 0.3268 & 0.2500
& 0.5138 \\
& LoRAP
& 9.7889 & 45.4406
& \textbf{0.7015} & 0.7012 & \textbf{0.5135} & 0.6393
& 0.6061 & \textbf{0.3788} & 0.2520
& 0.5418 \\
& CASP
& \textbf{9.7889} & \textbf{39.0195}
& 0.6798 & 0.7127 & 0.4940 & \textbf{0.6440}
& 0.6574 & 0.3618 & \textbf{0.2680}
& \textbf{0.5454} \\
\midrule

\multirow{4}{*}{50\%}
& LLM-Pruner
& 63.5828 & 217.6335
& 0.4300 & 0.6415 & 0.3408 & 0.5249
& 0.4773 & 0.2867 & 0.2120
& 0.4162 \\
& FLAP
& 50.2000 & 179.0200
& 0.3780 & 0.6344 & 0.3397 & 0.5422
& 0.4790 & 0.2773 & 0.2020
& 0.4075 \\
& LoRAP
& 49.7121 & 199.7117
& 0.6416 & 0.6159 & 0.3692 & 0.5485
& 0.4402 & 0.2739 & 0.2040
& 0.4419 \\
& CASP
& \textbf{18.0397} & \textbf{67.6835}
& \textbf{0.6436} & \textbf{0.6529} & \textbf{0.3739} & \textbf{0.5525}
& \textbf{0.5602} & \textbf{0.2978} & \textbf{0.2220}
& \textbf{0.4718} \\
\bottomrule
\end{tabular}%
}
\label{tab:main_large_table_sheet_vicuna7b}
\end{table*}
% (Insert Vicuna results table here)

\subsection{Results on GQA models}
\label{appendix:gqa}

\textcolor{red}{We further evaluate the proposed pruning algorithm on recent models with grouped-query attention (GQA). In GQA, multiple query heads share the same key-value head; therefore, we prune query heads as the primary units and remove a key-value head only after all query heads associated with it are pruned.}

\textcolor{red}{As shown in Table~\ref{tab:main_large_table_sheet_llama3_8b}, the results on LLaMA3-8B demonstrate that CASP remains effective on GQA-based models. Under the relatively low pruning ratio of 10\%, CASP achieves the best perplexity on both WikiText2 and PTB, while maintaining zero-shot performance comparable to the dense model. Under the aggressive pruning ratio of 50\%, CASP exhibits the strongest performance retention among the compared methods, as reflected by both lower perplexity and higher average zero-shot accuracy.}

\begin{table*}[ht]
\centering
\scriptsize
\setlength{\tabcolsep}{3.2pt}
\renewcommand{\arraystretch}{1.15}
\caption{Zero-shot performance of the compressed LLaMA3-8B. The average score is computed across seven datasets. The "\textbf{bolded}" represents the best result under the same pruning ratio.}
\resizebox{\textwidth}{!}{%
\begin{tabular}{cc|cc|ccccccc|c}
\toprule
Ratio & Method
& WikiText2 $\downarrow$ & PTB $\downarrow$
& BoolQ $\uparrow$ & PIQA $\uparrow$ & HellaSwag $\uparrow$ & WinoGrande $\uparrow$
& ARC-e $\uparrow$ & ARC-c $\uparrow$ & OBQA $\uparrow$
& Average $\uparrow$ \\
\midrule

0\% & Ground Truth
& 6.1367 & 10.5469
& 0.8226 & 0.8063 & 0.7922 & 0.7364
& 0.7752 & 0.5410 & 0.4480
& 0.7031 \\
\midrule

\multirow{3}{*}{10\%}
& LLM-Pruner
& 8.7911 & 13.9107
& 0.6881 & 0.7671 & 0.7231 & 0.7151
& 0.6881 & 0.4437 & 0.4200
& 0.6350 \\
& LoRAP
& 7.5508 & 12.4922
& 0.7783 & 0.7780 & 0.7503 & 0.7017
& \textbf{0.7369} & \textbf{0.5009} & \textbf{0.4240}
& 0.6672 \\
& CASP
& \textbf{7.1484} & \textbf{12.0156}
& \textbf{0.8040} & \textbf{0.7878} & \textbf{0.7595} & \textbf{0.7261}
& 0.7353 & 0.5000 & 0.4140
& \textbf{0.6752} \\
\midrule

\multirow{3}{*}{50\%}
& LLM-Pruner
& 60.0247 & 94.5353
& 0.5180 & 0.5696 & 0.3351 & 0.5130
& 0.3729 & 0.2432 & 0.2900
& 0.4060 \\
& LoRAP
& 92.8750 & 188.3750
& 0.5138 & \textbf{0.6507} & \textbf{0.4229} & 0.5399
& 0.3906 & 0.2594 & 0.2980
& 0.4393 \\
& CASP
& \textbf{33.2500} & \textbf{53.0312}
& \textbf{0.6072} & 0.6170 & 0.3761 & \textbf{0.5414}
& \textbf{0.4221} & \textbf{0.2918} & \textbf{0.3080}
& \textbf{0.4520} \\
\bottomrule
\end{tabular}%
}
\label{tab:main_large_table_sheet_llama3_8b}
\end{table*}

\end{document}